\documentclass[letterpaper, 10 pt, conference]{ieeeconf}  

\IEEEoverridecommandlockouts                              

\usepackage[bookmarks=true]{hyperref}
\usepackage{color}
\usepackage{cite}
\usepackage{notoccite}

\usepackage{graphicx}
\def\argmin{\mathop{\arg\min}}	%
\def\argmax{\mathop{\arg\max}}	%
\usepackage{bm}
\usepackage{tabularx}
\usepackage{floatrow}
\usepackage[utf8]{inputenc}
\usepackage{amsmath}
\usepackage[utf8]{inputenc} 
\usepackage[T1]{fontenc}    
\usepackage[mathscr]{eucal}
\usepackage{caption}
\usepackage{amsfonts}
\usepackage{dblfloatfix} 
\usepackage{epsfig}
\usepackage{makecell}
\usepackage{multirow}
\usepackage{hyperref}
\usepackage{dsfont}
\usepackage{breqn}
\usepackage{amssymb}
\usepackage{amsmath,amsthm}
\usepackage{amsfonts}       
\usepackage{nicefrac}       
\newtheorem{theorem}{Theorem}
\usepackage{amsthm}
\usepackage{bbm}

\newtheorem{proposition}{Proposition}

\input{mysymbol.sty}

\usepackage{mathtools,algorithm}
\usepackage[noend]{algpseudocode}
\title{\LARGE \bf Large Scale Distributed Collaborative Unlabeled Motion Planning with Graph Policy Gradients
}

\author{Arbaaz Khan$^{1}$, Vijay Kumar$^{1}$, Alejandro Ribeiro$^{1}$ 
\thanks{$^{1}$GRASP Lab, University of Pennsylvania, USA
        {\tt\small arbaazk@seas.upenn.edu}}%
}

\begin{document}

\maketitle
\thispagestyle{empty}
\pagestyle{empty}

\begin{abstract}
In this paper, we present a learning method to solve the unlabelled motion problem with motion constraints and space constraints in 2D space for a large number of robots. To solve the problem of arbitrary dynamics and constraints we propose formulating the problem as a multi-agent problem. We are able to demonstrate the scalability of our methods for a large number of robots by employing a graph neural network (GNN) to parameterize policies for the robots. The GNN reduces the dimensionality of the problem by learning filters that aggregate information among robots locally, similar to how a convolutional neural network is able to learn local features in an image. Additionally, by employing a GNN we are also able to overcome the computational overhead of training policies for a large number of robots by first training graph filters for a small number of robots followed by zero-shot policy transfer to a larger number of robots. We demonstrate the effectiveness of our framework through various simulations. 
\end{abstract}

\section{Introduction}
\label{sec:Sec1}
In robotics, one is often tasked with designing algorithms to coordinate teams of robots to achieve a task. 
In this paper, we concern ourselves with scenarios where a team of homogeneous or identical robots must execute a set of identical tasks such that each robot executes only one task, but it does not matter which robot executes which task. Concretely, this paper studies the concurrent goal assignment and trajectory planning problem where robots must simultaneously assign goals and plan motion primitives to reach assigned goals. This is the unlabelled multi-robot planning problem where one must simultaneously solve for goal assignment and trajectory optimization. 

In the past, several methods have been proposed to achieve polynomial-time solutions for the unlabelled motion planning problem \cite{adler2015efficient,yu2012distance,turpin2014capt}. The common theme among these methods is the design of a heuristic best suited for the robots and the environment. 
However, when additional constraints such as constraints on dynamics, presence of obstacles in the space, desired goal orientations; solving for a simple heuristic is no longer the optimal solution. In contrast to previous literature that relies on carefully designed but ultimately brittle heuristic functions, instead we hypothesize using reinforcement learning (RL) to compute an approximate solution for the assignment and planning problem without relaxing any constraints. When using RL to learn policies for the robots, one can optimize a cost function that is independent of any dynamics or constraints, i.e assign a high value if all goals are covered and zero otherwise. 

To achieve scalable multi-robot motion planning, we look to exploit the inherent graph structure among the robots and learn policies using local features only. We hypothesize that graph neural networks (GNNs) \cite{kipf2016semi,wu2019comprehensive} can be a good candidate to parameterize policies for robots as opposed to the fully connected networks used in \cite{khan2019learning}. GNNs work similar to convolutional neural networks (CNNs) and can be seen as an extension of CNNs to data placed on irregular graphs instead of a two dimensional grid such as an image. 
We define a graph $\mathcal{G} = (\mathbf{V},\mathbf{E})$  where $\mathbf{V}$ is the set of nodes representing the robots and $\mathbf{E}$ is the set of edges defining relationships between them. These relationships can be arbitrary and are user defined. In this work we define edges between robots based on the Euclidean distance between them. This graph acts as a support for the data vector $\bbX=[\bbx_1,\ldots,\bbx_N]^\top$ where $\bbx_n$ is the state representation of robot $n$. The GNN consists of multiple layers of graph filters and at each layer the graph filters extract local information from a node's neighbors. The information is propagated forward between layers after passing it through a non linear activation function similar to how one would propagate information in a CNN. The output of the final layer is given by $\Pi = [\pi_1,\ldots,\pi_N]$,  where $\pi_1,\ldots,\pi_N$ are independent control policies for the robots. During training, each robot rolls out its own trajectory by executing its respective policy. Each robot also collects a centralized reward and using policy gradient methods ~\cite{sutton1998reinforcement} the weights of the GNN are updated. Further, since the robots are homogeneous and the graph filters only learn local information, we circumvent the computational burden of training many robots by training the GNN on only a small number of robots but during inference use the same filter across all robots. 

\subsection{Related Work} 
The unlabeled motion planning is a relatively well studied problem in robotics \cite{adler2015efficient,yu2012distance,turpin2014capt,solovey2016hardness}. Concurrent assignment and planning (CAPT) \cite{turpin2014capt} is a provably correct and complete algorithm but requires centralized information. Further, CAPT makes assumptions about the dynamics and assumes the environment to be obstacle free. The work of \cite{solovey2015motion} proposes an algorithm for unlabeled motion planning for disk like obstacles. The algorithm has a guaranteed solution with a provable bound on runtime. It however requires the need to produce shortest paths for a robot moving in an environment with static obstacles which can prove to be a challenging problem in 3D. A recently emerging field of research looks to approximate intractable multi-robot problems with a model-free approach \cite{faust2018prm,khan2019learning} that provide significant benefits over analytic model based methods.  In our earlier work, \cite{khan2019learning}, we proposed the use of RL to solve the unlabeled motion planning problem with constraints on obstacles, dynamics of robots etc. However, a key feature lacking from \cite{khan2019learning} was scalability. We were only able to show unlabeled motion planning for 3-5 robots and this can mostly be attributed to the sample inefficiency of online deep RL methods. In another somewhat unrelated work, \cite{khan2019graph} we look to evaluate the use of a GNN to design decentralized policies for consensus in robots and are able to achieve results for a large number of robots. In this work, we look to \textit{\textbf{combine}} the results from both, i.e using RL for the unlabeled motion planning problem and achieve large scalability by employing graph convolutional networks to parameterize the decentralized policies. Through various simulations we demonstrate that the methods proposed in this paper provide a suitable solution that extends the learning formulation of the unlabeled motion planning algorithm to have the ability to scale to many robots.

%
\section{Distributed Collaborative Unlabeled Motion Planning} \label{sec:problem_formulation}

We consider the problem of navigating $N$ disk-shaped robots to a set of $N$ goals of radius $R$. We say that the navigation problem is unlabeled because there is no prior matching between robots and goals. Any robot can cover any goal. The positions of the goals are fixed and we use $\bbg_n$ to denote the position of goal $n$. The positions of robots change over discrete time index $t$ and are denoted as $\bbr_n(t)$ for robot $n$. We stack goal positions into the matrix $\bbG = [\bbg_1^T;\ldots;\bbg_N^T]$ whose $n$th row is the position of goal $n$ and robot positions into the matrix $\bbR(t) = [\bbr_1^T(t);\ldots;\bbr_N^T(t)]$ whose $n$th row is the position of robot $n$. We further group these two matrices to define the system state
\begin{equation}\label{eqn_system_state}
   \ccalS(t) \!= \langle \bbR(t), \bbG            \rangle   
             \!= \langle [\bbr_1^T(t);\ldots;\bbr_N(t)^T],
                     [\bbg_1^T;\ldots;\bbg_N^T]\rangle
\end{equation}
We assume that robot positions evolve according to a choice of action $\bba_{n}(t) \in \ccalA$ as determined by some stationary distribution with conditional density $p(\bbr_{n}(t+1) | \bbr_{n}(t),\ccalS_{n}(t))$. This first order dynamical model implies that the state $\ccalS(t)$ is a complete description of the system at time $t$. We further assume operation in open space. Both these choices are for simplicity of exposition. Incorporating obstacles (Section \ref{sec:exps}) and high order complex dynamics is an easy extension.   

The goal of the team of robots is to cover all goals in the sense that all goals $\bbg_n$ have at least one robot within distance $R$ of its location. To measure how well the configuration $\ccalS(t)$ in \eqref{eqn_system_state} accomplish this task, the reward $r(\ccalS(t))$ counts the number of goals covered. This is equivalent to counting all the goals for which $ \|\bbg_n - \bbr_m\| \leq R$ for at least one robot $m$, which we can write as the sum 
\begin{align} \label{eqn_reward}
   r(\ccalS(t)) 
      = \sum_{n=1}^N 
            \mbI\Big[ \min_m \|\bbg_n - \bbr_m (t)\| \leq R
                \Big] 
\end{align}
In \eqref{eqn_reward} only the robots that are within distance $R$ of a goal contribute to the reward. If there are several robots within $R$ of a goal, only the closest contributes toward increasing $r(\ccalS(t))$. The maximum reward $r(\ccalS(t))=N$ occurs when all goals are covered by one robot. Therefore, the necessary and sufficient condition for all goals to be covered at some time $t$ is for the reward to attain its maximum $r(\ccalS(t)) = N$.

Our interest here is designing policies to reach the maximum reward $r(\ccalS(t))=N$ (Section \ref{sec_gpg}) with a distributed collaborative system. The system is distributed because agents have access to local state information only and collaborative because nearby agents communicate with each other. To describe the locality of information and communication, consider orderings of goals and robots relative to their distance to a given robot. Formally, let $\bbr_{n[m]}(t) $ be the $m$th closest robot to robot $n$ other than $n$ itself so that for all $1\leq m\leq m' \leq N-1$ it holds
\begin{equation}\label{eqn_ordered_robot_positions}
   \| \bbr_{n[m]}(t)  - \bbr_n(t)  \| 
      \ \leq\  \| \bbr_{n[m']} (t) -  \bbr_n(t) \| .
\end{equation}
Likewise, let $\bbg_{n[m]}$ be the $m$th closest goal to robot $n$ so that for any $1\leq m\leq m' \leq N$ we can write
\begin{equation}\label{eqn_ordered_goal_positions}
   \| \bbg_{n[m]}(t)  - \bbr_n(t)  \| 
      \ \leq\  \| \bbg_{n[m']} (t) -  \bbr_n(t) \| .
\end{equation}
Observe that the position $\bbg_{n[m]}(t)$ of the $m$th closest goal changes over time as robot $n$ moves through the environment.

We assume that each agent senses \textit{at most} the $M$ goals and at most the $M$ robots that are closest to its location and that it can communicate with the $M$ closest robots. We therefore define the state of robot $n$ at time $t$ as a row vector concatenating its own location, the location of its $M$th closest robots, and the location of its $M$th closest goals,
\begin{align} \label{eq:robotstate}
    \bbx_n(t) = 
       \Big[\bbr_n^T(t); \
            \bbr_{n[1]}^T(t);\ldots; &\bbr_{n[M]}^T(t);  \nonumber \\
            &\bbg_{n[1]}^T(t);\ldots; \bbg_{n[M]}^T(t) \Big].
\end{align}
For future reference we introduce $d$ to denote the number of entries of $\bbx_n(t)$. This number is $d=2(2M+1)$ for planar navigation and $d=3(2M+1)$ for navigation in three dimensions. Communication between agents is also restricted to local information exchanges. We model this with a communication graph 
whose adjacency matrix we denote by $\bbS(t)\in\reals^{N\times N}$. Entries of this matrix $S_{nm} (t)$ are binary and are 1 only if robot $m$ is among the $M$th closest robots to robot $n$. We write this formally as 
\begin{align} 
\label{eq:graphstate}
   S_{nm} (t) 
     = \mbI \Big[ \bbr_m \in 
                    \big\{ \bbr_n(t), \bbr_{n[1]}(t), \ldots, &\bbr_{n[M]}(t) \big\}
            \Big]
\end{align}
where we point out that we also add self loops to $\bbS$ because we have made $S_{nn}=1$. Observe that $M$ limits the number of robots that communicate with each robot $n$, as well as the number of robots and goals sensed by each robot $n$. These three parameters can be distinct in practical implementations but we make them equal here to simplify notation. 

Robots are given access to their local states $\bbx_n(t)$ as defined in \eqref{eq:robotstate} and can exchange information with neighboring agents as dictated by the graph $\bbS(t)$ whose entries are given by \eqref{eq:graphstate}. Given their local states and information received from neighbors, robot $n$ chooses an $e$-dimensional action $\bba_{n}(t) \in \reals^e$ that controls the transition into position $\bbr_n(t+1)$ according to the dynamical model $p(\bbr_{n}(t+1) | \bbr_{n}(t),\bba_{n}(t))$. We write these stochastic policies as:
\begin{align} \label{eqn_policies}
   \bba_n(t) = \pi_n \big(\bba_{n}(t) \given \bbx_{n}(t); \bbS(t) \big).
\end{align}
The notation in \eqref{eqn_policies} signifies that $\bbS_{n}(t)$ is chosen according to the local state $\bbx_{n}(t)$ -- i.e., the policy is distributed -- and information exchanges with neighboring nodes as dictated by the graph $\bbS(t)$ -- i.e., the policy is collaborative. We emphasize that the graph $\bbS(t)$ affects the choice of action because it limits the information accessible to robot $n$ and that it therefore is an important component of the robots policy. Notwithstanding, these graph is not necessarily known to robot $n$ and our policies will be executable without this knowledge (Section \ref{sec_gpg}). The individual distributed collaborative policies in \eqref{eqn_policies} generate the product policy $\Pi = \pi \big(\bba_{1}(t) \given \bbx_{1}(t); \bbS(t) \big)\times\ldots\times\pi \big(\bba_{N}(t) \given \bbx_{N}(t); \bbS(t) \big)$. We want to jointly optimize over individual policies to maximize the discounted reward 
\begin{equation} \label{eq:cost_function_prel}
   J({\Pi}) = \max_{\Pi}  \mathbb{E}_{\Pi}\bigg[\sum_{t}^T \gamma^t r(\ccalS(t))\bigg] 
\end{equation} 
where the reward $r(\ccalS(t))$ is as given in \eqref{eqn_reward}. Our objective in this paper is to learn policies having the form in \eqref{eqn_policies} that maximize the discounted reward in \eqref{eq:cost_function_prel}. We do so with the Graph Policy Gradient method that we introduce next.

%

%
\section{Graph Policy Gradient}\label{sec_gpg}

To find the policy $\Pi$ that maximizes the reward $J({\Pi})$ in \eqref{eq:cost_function} it is customary to introduce a policy parameterization. Our key technical innovation is to parameterize $\Pi$ with a graph neural network (GNN), which as we will explain in Section \ref{sec_invariance} not only respects but also leverages the locality of the distributed collaborative motion planning problem we discussed in Section \ref{sec:problem_formulation}. GNNs are generalizations of convolutional neural networks (CNNs) built on the definition of convolutional graph filters.
Formally, define the state matrix $\bbX(t)\in\reals^{n\times d}$ whose $n$th row is the state of robot $n$ [cf. \eqref{eq:robotstate}],
\begin{align} \label{eqn_all_robots_state}
    \bbX(t) = [\bbx_1(t); \ldots; \bbx_N(t) ]  \in \reals^{N\times d} .
\end{align}
A graph convolutional filter to process $\bbX(t)$ on the graph $\bbS(t)$ is defined by a set of $K$ coefficient matrices $\bbH_{1k} \in \reals^{d\times d_1}$. These matrices serve as coefficients of a polynomial on $\bbS(t)$ that operates on the state $\bbX(t)$ to produce the output $\bbY(t)\in\reals^{d_1}$ given by
\begin{equation}\label{eqn_graph_filter}
   \bbZ_1(t) = \sum_{k=0}^{K-1} \bbS^k(t) \bbX(t) \bbH_{1k} .
\end{equation}
The output of the graph filter in \ref{eqn_graph_filter} is further processed with a pointwise nonlinearity to produce the layer 1 output signal  
\begin{equation}\label{eqn_layer_1}
   \bbX_1(t) = \sigma\Big[\ \bbZ_1(t) \ \Big] 
             = \sigma\bigg[ \sum_{k=0}^{K-1} \bbS^k(t) \bbX(t) \bbH_{1k} \bigg].
\end{equation}
In a GNN with multiple layers the signal $\bbX_1(t)$ is further processed with a graph filter with coefficients $\bbH_{1k} \in \reals^{d_1\times d_2}$ and a pointwise nonlinearity, to produce the output of Layer 2. In general, there are a total of $L$ layers each of which is defined by a set of $K$ coefficients $\bbH_{1k} \in \reals^{d_{l-1}\times d_l}$ which produce output signals $\bbX_l(t)$ according to the recursion 
\begin{equation}\label{eqn_gnn_recursion}
   \bbX_l(t) = \sigma\Big[\ \bbZ_l(t) \ \Big] 
             = \sigma\bigg[ \sum_{k=0}^{K-1} \bbS^k(t) \bbX_{l-1}(t) \bbH_{lk} \bigg].
\end{equation}
The output of the $L$th layer is the output of the GNN and we propose here to use it as the mechanism for generating robot actions. Specifically, stack the individual actions of all agent into the action matrix $\bbA(t) = [\bba_1^T(t); \ldots; \bba_N^T(t)]\in\reals^{n\times e}$ and make 
\begin{equation}\label{eqn_gnn_output}
   \bbA(t) = \bbX_L(t)
           = \Phi (\bbX(t), \bbS(t); \bbH).
\end{equation}
Notice that in \eqref{eqn_gnn_output} we have introduced the notation $ \Phi (\bbX(t), \bbS(t); \bbH)$ to represent the GNN's output. This output depends on the state input $\bbX(t)$, the graph $\bbS(t)$ and the filter tensor $\bbH:=\{\bbH_{lk}\}_{l,k}$ that groups the coefficient matrices of all layers and all orders. The advantage of using a GNN to parameterize actions $\bbA(t)$ is that they respect the structure of the distributed collaborative policies in \eqref{eqn_policies}. The graph filters in \eqref{eqn_graph_filter} and \eqref{eqn_gnn_recursion} are made up of diffusion operations that involve interactions between neighbors only. Consider the product $\bbS(t)\bbX(t)$ and observe that the sparsity pattern of $\bbS(t)$ is such that the $n$th row of this product is given by 
\begin{equation}\label{eqn_locality}
   [ \bbS(t) \bbX(t) ]_n
           = \sum_{m: S_{mn}=1} \bbx_{m}(t)
\end{equation}
If we interpret the row $ \big[ \bbA(t) \bbX(t) \big]_n$ as a quantity that is evaluated by robot $n$, it follows that robot $n$ can evaluate this row by communicating with neighboring nodes only. Subsequent entries in the graph filters in \eqref{eqn_graph_filter} can be recursively evaluated noticing that $ \bbS^k(t) \bbX(t) = \bbS^{k-1}(t) \bbX(t)$ and that we can therefore write:
\begin{equation}\label{eqn_more_locality}
   [ \bbS^k(t) \bbX(t) ]_n
           = \sum_{m: S_{mn}=1} [ \bbS^{k-1}(t) \bbX(t) ]_m
\end{equation}
It follows that all of the summands in \eqref{eqn_graph_filter} can be computed exclusively through local information exchanges. The {\it pointwise} nonlinearity in \eqref{eqn_layer_1} can also be locally implemented. Subsequent layers can be evaluated in a distributed manner as well, because the argument in \eqref{eqn_locality} and \eqref{eqn_more_locality} is not specific to Layer 1. Through this recursive distributed computations, robot $n$ ends up computing the $n$th row of the GNN output $[\bbX_L(t)]_n = [\Phi (\bbX(t), \bbA(t); \bbH)]_n$. This row is used to define the local policy in \eqref{eqn_policies} as
\begin{align} \label{eqn_policies_from_gnn}
   \bba_n(t) = [\bbA(t)]_n 
             = [\bbX_L(t)]_n 
             = [\Phi (\bbX(t), \bbS(t); \bbH)]_n
\end{align}
The graph policy gradient method (GPG) is the search for a joint policy $\Pi$ that maximizes the cost in \eqref{eq:cost_function} over the space of policies $\Pi(\bbH)$ that produce actions parameterized with a GNN of the form in \eqref{eqn_gnn_recursion}. The solution of this optimization is the optimal filter tensor
\begin{align} \label{eq:cost_function}
   \bbH^* = \argmax_{\bbH} \
                & \mathbb{E}_{\Pi(\bbH)} \bigg[\sum_{t}^T \gamma^t r(\ccalS(t))\bigg] \nonumber \\
                & \Pi(\bbH) : \bbA(t) = \Phi (\bbX(t), \bbS(t); \bbH) .    
\end{align} 
In \eqref{eq:cost_function}, the filter tensor $\bbH$ along with the state $\bbX(t)$ and the graph $\bbS(t)$ determine the choice of action $\bbA(t)$. This joint action controls the state transition as dictated by the dynamical model $p(\bbr_{n}(t+1) | \bbr_{n}(t),\bbS_{n}(t))$. We want to find a filter tensor $\bbH$ that results in the maximum reward $J(\Pi(\bbH))$. 

The optimal filter can be computed through online policy gradient methods. Robots roll out trajectories $\tau \sim \Pi$ and collect rewards $r(\mathcal{S}(t))$ over a time horizon $t=0,\ldots,T$. Since the policies are assumed to be independent, the policy gradient wrt $\mathbf{H}$, $\nabla_{\mathbf{H}}J$ can be given as:
\begin{equation}
\label{eq:policygradient}
     = \mathbb{E}_{\tau \sim (\Pi)}\Bigg[\Big(\sum_{t=1}^T\nabla_{\mathbf{H}} \log[\pi_1 \times \ldots \times \pi_N]\Big) \Big(\sum_{t=1}^T r(\mathcal{S}(t)) \Big) \Bigg] 
\end{equation}

This is called Graph Policy Gradients or GPG and was first proposed in \cite{khan2019graph} as a method of taking gradients for RL policies parameterized by graph neural networks. In parameterizing actions with a GNN we ensure the possibility of having a distributed implementation. We will see in the upcoming Section \ref{sec_invariance} that GNNs also exhibit an invariance to the labeling that substantiates transferability properties that we explore in the numerical experiments in Section \ref{sec_experiments}.

%
\section{Permutation Invariance of GNN Policy parameterizations}\label{sec_invariance}

Let the GNN coefficients $\mathbf{H}^*$ be the solution that maximizes (\ref{eq:cost_function}) for a given $\bbS(t)$ and $\bbX(t)$. Consider another set of graphs $\tilde{\bbS}(t)$ and state vectors $\tilde{\bbX}(t)$ that are produced by permuting $\bbS(t)$ and $\bbX(t)$. Let the filter coefficients after training on $\tilde{\bbS}(t)$ and $\tilde{\bbX(t)}$ be $\tilde{\mathbf{H}}^*$. To study the relationship between $\mathbf{H}^*$ and $\tilde{\mathbf{H}}^*$ we first define a set of permutation matrices of dimension ${N}$ such that $\bm{\mathcal{P}} = \{\mathbf{{P}}\in \{0,1\}^{{N} \times {N}}\    \mathbf{{P1=1}},\mathbf{{P}^\top1 =1}\}$


Such a  permutation matrix $\mathbf{P}$ is one for which the product $\mathbf{P}^{\top}\bbX(t)$ reorders the entries of any $\bbX(t)$ and the operation $\mathbf{P}^{\top}\bbS(t)\mathbf{P}$ produces a reordering of the rows and columns of any given $\bbS(t)$. The policy for system with $\bbS(t)$ is $\Pi = \Phi(\bbX(t),\bbS(t);\mathbf{H})$. Analogously, let the policy for the permuted system be given as:
\begin{align}
\begin{split}
\tilde{\Pi}=\Phi({\tilde{\bbX}(t),\tilde{\bbS}(t)};\tilde{\mathbf{H}}) \text{ where }     
  \tilde{\bbS}(t)=\mathbf{P}^{\top}\bbS(t)\mathbf{P},\\
\end{split}
\end{align}
$\tilde{\bbX}$ is the permuted state of all robots and $\tilde{\mathbf{H}}$ is the vector of filter coefficients that parameterizes $\tilde{\Pi}$. This leads to:
\begin{theorem}
\label{thm:permutationinvariance}
Given a configuration of robots represented by $\bbX(t)$ that define an underlying graph $\mathcal{G}$  with graph shift operator $\bbS(t)$ 
and another configuration of robots given by $\tilde{\bbX}(t)$ and $\tilde{\bbS}(t)$ 
then the graph filter coefficients $\mathbf{H}^*$ and $\tilde{\mathbf{H}}^*$ which are the optimal solutions for the systems $(\bbS(t),\bbX(t))$ and $(\tilde{\bbS}(t),\tilde{\bbX}(t))$ respectively are equivalent,
\begin{equation}
    \mathbf{H}^* \equiv \tilde{\mathbf{H}}^*  
\end{equation}
\end{theorem}
In order to prove Theorem \ref{thm:permutationinvariance}, we must prove that both the unlabeled motion planning problem and the GNNs parameterized by $\tilde{\mathbf{H}}^* \text{ and } \mathbf{H}^*$ are permutation equivariant. 

\subsection{Equivariance of GNNs for Unlabeled Motion Planning}
\begin{proposition}
\label{prop:propstate} Given a system of robots and goals $ \ccalS(t) = \{ \bbr(t),\, \bbg\}   $ and another system where the robot states are permuted $\tilde{\mathcal{S}}(t)= (\tilde{\mathbf{r}}(t) ,\mathbf{g})$ where $\tilde{\mathbf{r}}(t)=\mathbf{P}^\top\mathbf{r}(t)$, the corresponding robot states $\bbX(t)=[\bbx_1(t),\ldots,\bbx_{{N}}(t)]^\top$ and $\tilde{\bbX}(t)=[\tilde{\bbx}_1(t),\ldots,\tilde{\bbx}_{{N}}(t)]^\top$ are permutation related, i.e $\tilde{\bbX}(t)=\mathbf{P}^\top\bbX(t).$   
\end{proposition}
\textbf{Proof} This is true because we have used ordering to construct the local robot states in \eqref{eq:robotstate} and orderings are invariant to permutations. Formally, consider indexes $n$ and $\tdn$ with $[\bbP]_{n\tdn}=1$. It follows from $\tbR(t) = \bbP^T\bbR(t)$ that $\bbr_n(t) = \bbr_\tdn(t)$; which means that robot $n$ has been mapped to robot $\tdn$ in the permutation. Since orderings are independent of labeling it follows that for all robots we must have $\bbr_{n[m]}(t) = \bbr_{\tdn[m]}(t) $ [cf. \eqref{eqn_ordered_robot_positions}]. Likewise, for all goals we must have $\bbg_{n[m]}(t) = \bbg_{\tdn[m]}(t)$ [cf. \eqref{eqn_ordered_goal_positions}]. Given the definition of the state row vector in \eqref{eq:robotstate} we therefore have that $\bbx_{n}(t) = \bbx_{\tdn}(t)$. Further recalling the definition of the state matrix $\bbX(t)$ in \eqref{eqn_all_robots_state} and the assumption that $[\bbP]_{n\tdn}=1$ we also have that $\bbx_n(t) = [\bbP^T\bbX(t)]_\tdn$. Putting the latter two statements together we conclude that $[\bbP^T\bbX(t)]_\tdn=  \bbx_{\tdn}(t)$. Since this is true for arbitrary $\tdn$ we must have $\bbP^T\bbX(t)=\tilde{\bbX}(t)$. In order to show that the GNN parameterization is equivariant for both settings, consider the following proposition: 
\begin{proposition}
\label{prop:no2}
Given robots with states $\bbX(t)$ and $\tilde{\bbX}(t)$ and underlying graphs $\bbS(t)$ and $\tilde{\bbS}(t)$ such that $\tilde{\bbX}(t)=\mathbf{P}^{\top}\bbX(t)$ (from Proposition \ref{prop:propstate}) and $\tilde{\bbS}(t)=\mathbf{P}^\top\bbS(t)\mathbf{P}$ for some permutation matrix $\mathbf{P}$, the outputs of a GNN policy $\Phi$ with filter coefficients $\mathbf{H}$ to the pairs $(\bbS(t),\bbX(t))$ and $(\tilde{\bbS}(t),\tilde{\bbX}(t))$ are related by: 
\begin{equation}
    \Phi(\tilde{\bbX}(t),\tilde{\bbS}(t);\mathbf{H})=\mathbf{P}^{\top}\Phi(\bbX(t),\bbS(t);\mathbf{H})
\end{equation}
\end{proposition}

\textbf{Proof }The output of the GNN filter for the system $(\tilde{\bbS}(t),\tilde{\bbX}(t))$ as given in (\ref{eqn_gnn_recursion}) is:
\begin{equation}
\Phi(\tilde{\bbX}(t),\tilde{\bbS}(t);\mathbf{H}) = \sum_{k=0}^{K} h_k \tilde{\bbS}(t)^k \tilde{\bbX}(t)   
\end{equation}
This can be expressed as: 
\begin{equation}
    \sum_{k=0}^{K} h_k \tilde{\bbS}(t)^k \tilde{\bbX} = \sum_{k=0}^\infty h_k (\mathbf{P}^\top\bbS(t)^k\mathbf{P})\mathbf{P}^\top \bbX(t)
\end{equation}
Using the fact that $\mathbf{P}$ is an orthogonal matrix which in turn implies that $\mathbf{P}^\top \mathbf{P} = \mathbf{P}\mathbf{P}^\top = \mathbf{I}$:
\begin{align}
    \sum_{k=0}^\infty h_k (\mathbf{P}^\top\bbS(t)^k\mathbf{P})\mathbf{P}^\top \bbX(t) = \mathbf{P}^\top \sum_{k=0}^\infty h_k \bbS(t)^k \bbX(t)  \nonumber \\
    =\mathbf{P}^\top\Phi(\bbX(t),\bbS(t);\mathbf{H})
\end{align}
Thus, proving Proposition \ref{prop:no2}. Intuitively, Proposition \ref{prop:no2} tells us that reordering the robots states and corresponding nodes in the graph representation which is fed into the GNN policy, results in an appropriate reordering of the outputs of the filter without any change in the weights of the policy.

\subsection{Equivariance of Unlabeled Motion Planning} 
In order to show that the unlabeled motion planning is permutation equivariant, we look at the cost $J$. 

\begin{proposition}
\label{prop:propno1}
The reward function $J$ is permutation equivariant. i.e for all permutation matrices $\mathbf{P} \in \mathcal{P}$, $J_{\Pi}= J_{\tilde{\Pi}}$ 
\end{proposition}

\textbf{Proof} For the permuted system  the reward is given as : 
\begin{align} 
\label{eq:permutedrewardstruc}
   r(\tilde{\ccalS}(t)) 
      = \sum_{n=1}^N 
            \mbI\Big[ \min_m \|\bbg_n - \tilde{\bbr}_m\| \leq R
                \Big] 
\end{align}

Since the set of robots $\mathbf{r}$ and $\tilde{\mathbf{r}}$  are the same:
\begin{equation}
    \label{eq:pidefinition}
    \tilde{\mathbf{r}}_{i[1]} = \argmin_j \text{ } \lvert \lvert \mathbf{g}_i - \tilde{\mathbf{r}}_j \rvert\rvert =    {\mathbf{r}}_{i[1]}
\end{equation}
Which leads us to $r(\ccalS(t))=r(\tilde{\ccalS}(t))$ and therefore $\sum_{t}^T r(\ccalS(t)) = \sum_{t}^T r(\tilde{\ccalS}(t))$.   
Now, 
\begin{align}
J_{\tilde{\Pi}} = \sum_{n=1}^{{N}} \max_{\theta}  \mathbb{E}_{\tilde{\Pi}}\bigg[\sum_{t}^T r(\tilde{\ccalS}(t))\bigg]  \\= \sum_{n=1}^{{N}} \max_{\theta} \int \sum_{t}^T r(\tilde{\ccalS}(t)) d\tilde{\Pi}
\end{align}

Using the fact that $d\tilde{\Pi} = \mathbf{P}^{\top}d{\Pi}$ (from Proposition \ref{prop:no2}) and $r(\ccalS(t))=r(\tilde{\ccalS}(t))$ we get; 
\begin{equation}
  \sum_{n=1}^{{N}} \max_{\theta}  \int \sum_{t}^T r(\tilde{\ccalS}(t)) d\tilde{\Pi} = \sum_{n=1}^{\mathbf{N}} \max_{\theta} \mathbf{P}^\top \int \sum_{t}^T r(\ccalS(t)) d{\Pi}
\end{equation}
Since the set of robots is $\mathbf{N}$ for both $J_{\Pi}$ and $J_{\tilde{\Pi}}$, we can conclude $J_{\tilde{\Pi}}=J_{\Pi}$
Thus, we conclude that the unlabeled motion planning problem is permutation equivariant. With these results, we construct the following proof: \\
\textbf{Proof for Theorem \ref{thm:permutationinvariance}} Let optimal coefficients $\mathbf{H}^*$ and $\tilde{\mathbf{H}}^*$ induce a optimal reward $J^*$ and $J_{\tilde{\Pi}}^*$ respectively. Consider the optimal filter coefficient $\mathbf{H}$. From proposition \ref{prop:no2}, we have:
\begin{equation}
\Phi(\tilde{\bbX}(t),\tilde{\bbS}(t);\mathbf{H})=\mathbf{P}^{\top}\Phi(\bbX(t),\bbS(t);\mathbf{H})
\end{equation}
$\Phi(\tilde{\bbX}(t),\tilde{\bbS}(t);\mathbf{H})$ induces a reward $J_{\tilde{\Pi}}$ while $\mathbf{P}^{\top}\Phi(\bbX(t),\bbS(t);\mathbf{H})$ induces the optimal reward $J_{\Pi}^*$. However, as a result of Proposition \ref{prop:propno1}, we have:
\begin{equation}
\label{eq:costequal1}
    J_{\tilde{\Pi}} = J_{\Pi}^*
\end{equation}
Similarly by considering the optimal filter coefficient $\tilde{\mathbf{H}}^*$: 
\begin{equation}
\label{eq:costequal2}
J_{\tilde{\Pi}}^* = J_{\Pi}    
\end{equation}
If the cost $J_{\Pi}^*$ and $J_{\tilde{\Pi}}^*$ are the optimal costs, then:
\begin{align}
\label{eq:inequals}
    J_{\Pi}^* \geq J_{\Pi} = J_{\tilde{\Pi}}\\    
\label{eq:inequals2}
    J_{\tilde{\Pi}}^* \geq J_{\tilde{\Pi}} = J_{\Pi}^*
\end{align}
where the second equality follows from (\ref{eq:costequal1}) and (\ref{eq:costequal2}). From (\ref{eq:inequals}) and (\ref{eq:inequals2}), we can conclude that 
\begin{equation}
    \label{eq:costequals}
    J_{\tilde{\Pi}}^* = J_{\Pi}^*
\end{equation}
Thus, we can conclude $ \mathbf{H}^* \equiv \tilde{\mathbf{H}}^*$, completing the proof for Theorem \ref{thm:permutationinvariance}.
One of the bottlenecks of RL for multi-robot systems is the fact that it lacks the ability to scale to a large number of robots. However, Theorem 1 and Proposition \ref{prop:no2} offer us a valuable tool. When training, we train only for a small number of robots with graph $\bbS(t)$ which yields graph filters $\mathbf{H}$. By realizing that the bigger swarm consists of smaller swarms that are permutations $\bbS(t)$ and the filter coefficients for both are equivalent, we simply reuse the filter $\mathbf{H}$ on each of the smaller swarms. Thus, we are able to achieve large scale unlabeled motion planning without the computational burden of having to train a very large number of robots. 

\section{Experiments} \label{sec_experiments}
\begin{figure*}[t]
    \centering
    \includegraphics[width=\linewidth]{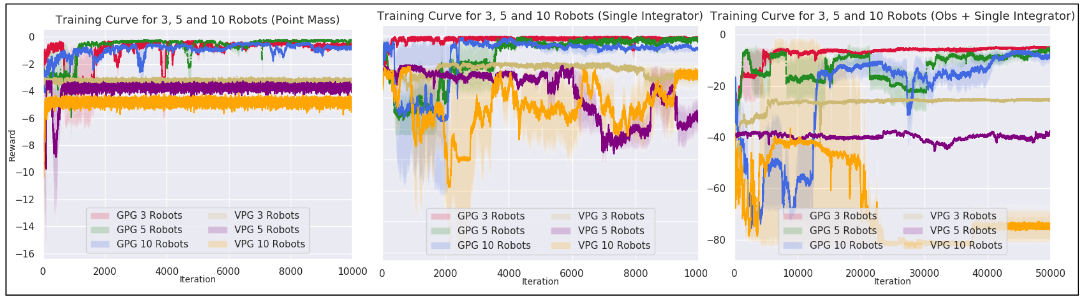}
    \caption{\textbf{Training Curves for 3, 5 and 10 robots} Policies trained by GPG are able to converge on experiments with point mass robots, experiments where robots follow single integrator dynamics and are velocity controlled as well as experiments when disk shaped obstacles are present in the environment.} 
    \label{fig:rew_curves}
\end{figure*}
\label{sec:exps}
To test the efficacy of GPG on the unlabelled motion planning problem, we setup a few experiments in simulation. We establish five main experiments. \textbf{1)} Unlabelled motion planning with three, five and  ten robots where robots obey point mass dynamics. \textbf{2)} GPG is tested on three, five and ten robots but here robots obey single integrator dynamics. \textbf{3)} The robots follow single integrator dynamics and additionally the environment is populated with disk shaped obstacles. \textbf{4)} The performance of GPG is tested against a model based provably optimal centralized solution for the unlabelled motion planning problem. We demonstrate empirically that the performance of GPG is almost always within a small margin of that of the model based method but comes with the additional advantage of being decentralized. \textbf{5)} The learning formulation for the unlabeled motion planning problem is agnostic to underlying robot dynamics or operating conditions in the environment. To demonstrate this, we look to execute the unlabeled motion planning problem in the higher order dynamics simulator AirSim \cite{shah2018airsim} equipped with quadrotors that model real world effects such as downwash and wind. 
To establish relevant baselines, we compare GPG with Vanilla Policy Gradients (VPG) where the policies for the robots are parameterized by fully connected networks (FCNs). Apart from the choice of policy parameterization there are no other significant differences between GPG and VPG.
\begin{figure*}[b!]
  \centering
  \includegraphics[width=\linewidth]{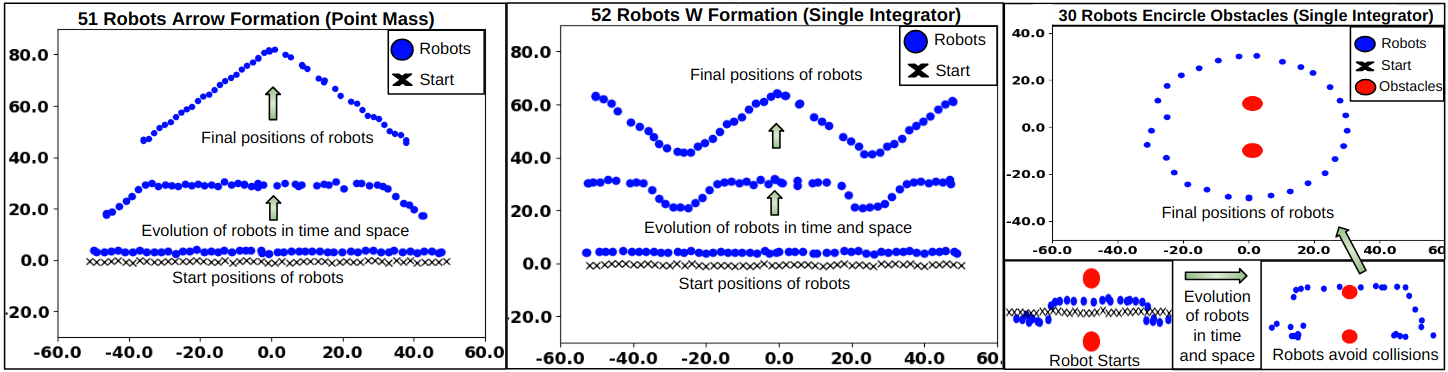}
  \caption{\textbf{Transferring Learned GPG Filters for Large Scale Unlabelled Motion Planning.} (Left) A small number of robots are trained to cover goals and follow point mass dynamics. During testing the number of robots as well as the distance of the goals from the start positions are much greater than those seen during training. (Center) A similar experiment is performed but now robots have single integrator dynamics. (Right) In this experiment in addition to single integrator dynamics, the environment also has obstacles that robots must avoid.\label{fig:formationfig}}
\end{figure*}
\subsection{Experimental Details and Training}
For GPG, we setup a L-layer GNN. For experiments involving 3 and 5 robots with point mass experiments  we find a 2 layer GNN, i,e a GNN that aggregates information from neighbors that are at most 2 hops away to be adequate. For experiments with 10 robots we find that GNNs with 4 layers work the best. For the baseline VPG experiments, we use with 2-4 layers of FCNs. The maximum episode length is $200$ steps and the discount factor $\gamma= 0.95$. In experiments with 3 robots, each robot senses 2 nearest goals and 1 nearest robot. In experiments with 5 and more  robots, robots sense 2 nearest goals and 2 nearest robots. The graph connects robots to their $1$,$2$ and $3$ nearest neighbors in experiments with $3,5$ and $10$ robots respectively. 

The behavior of GPG v/s VPG during training can be observed from Fig. \ref{fig:rew_curves}. We observe that in all cases GPG is able to produce policies that converge close to the maximum possible reward (in all three cases maximum possible reward is zero). When compared to the convergence plots of \cite{khan2019learning} who first proposed use of RL for the unlabelled motion planning problem, this represents a large improvement on just training.
It can also be observed that GPG converges when robot dynamics are changed or obstacles are added to the environment. While this is not necessarily the optimal solution to the unlabelled motion problem, it is an approximate solution to the unlabelled motion planning problem. The FCN policies represented by VPG in Fig. \ref{fig:rew_curves} fail to converge even on the simplest experiments. 

\subsection{Experimental Results - Inference}
The previous section shows the feasibility of GPG as a method for training a large swarm of robots to approximately solve the unlabelled motion planning problem. However, training a large number of robots is still a bottleneck due to the randomness in the system. Training 10 robots with simple dynamics on a state of the art NVIDIA 2080 Ti GPU with a 30 thread processor needs 7-8 hours. We see from our experiments that this time grows exponentially with an increase in the number of robots. 
Thus, to overcome this hurdle and to truly achieve large scale solutions for the unlabelled motion planning problem, we hypothesize that since the graph filters learned by GPG only operate on local information, in a larger swarm one can simply slide the same graph filter everywhere in the swarm to compute policies for all robots without any extra training. Intuitively, this can be attributed to the fact that while the topology of the graph does not change from training time to inference time, the size of the filter remains the same. As stated before, this is akin to sliding a CNN filter on a larger image to extract local features after training it on small images.  
To demonstrate the effect of GPG during inference time, we setup three simple experiments where we distribute goals along interesting formations. As described earlier, each robot only sees a certain number of closest goals, closest robots and if present closest obstacles. Our results can be seen in Fig. \ref{fig:formationfig}. The policies in Fig. \ref{fig:formationfig}(Left) and Fig. \ref{fig:formationfig} (Center) are produced by transferring policies trained to utilize information from 3-hop neighbors. In Fig. \ref{fig:formationfig} the policies are transferred after being trained to utilize information from 5-hop neighbors. Consider the formation shown in Fig \ref{fig:formationfig} (Left). Here each robot receives information about 3 of its nearest goals and these nearest goals overlap with its neighbors. Further, since the goals are very far away and robots are initialized close to each other, a robot and its neighbor receives almost identical information. In such a scenario the robots must communicate with each other and ensure that they each pick a control action such that they do not collide into each other and at the end of their trajectories, all goals must be covered. The GPG filters learn this local coordination and can be extended to every robot in the swarm. When training with obstacles, we extend the system state in Eqn. \ref{eqn_system_state} to include the positions of the obstacles $\bbO$ i.e $\mathcal S(t) \!= \{ \bbR(t), \bbG, \bbO \}$ and extend the state of the robots in Eqn. \ref{eq:robotstate} to observe $M$ nearest obstacles. The cost itself is unchanged, and as such all results derived in this paper still hold with the inclusion of the state of the obstacles. As before we train with the GNN filters with a small number of robots and extend the filters to a larger number. We observe in Fig \ref{fig:formationfig} (Right) and attached video with this paper that our solution is capable of covering goals even when obstacles are present.  

\begin{figure}[t!]
  \centering
  \includegraphics[width=\linewidth]{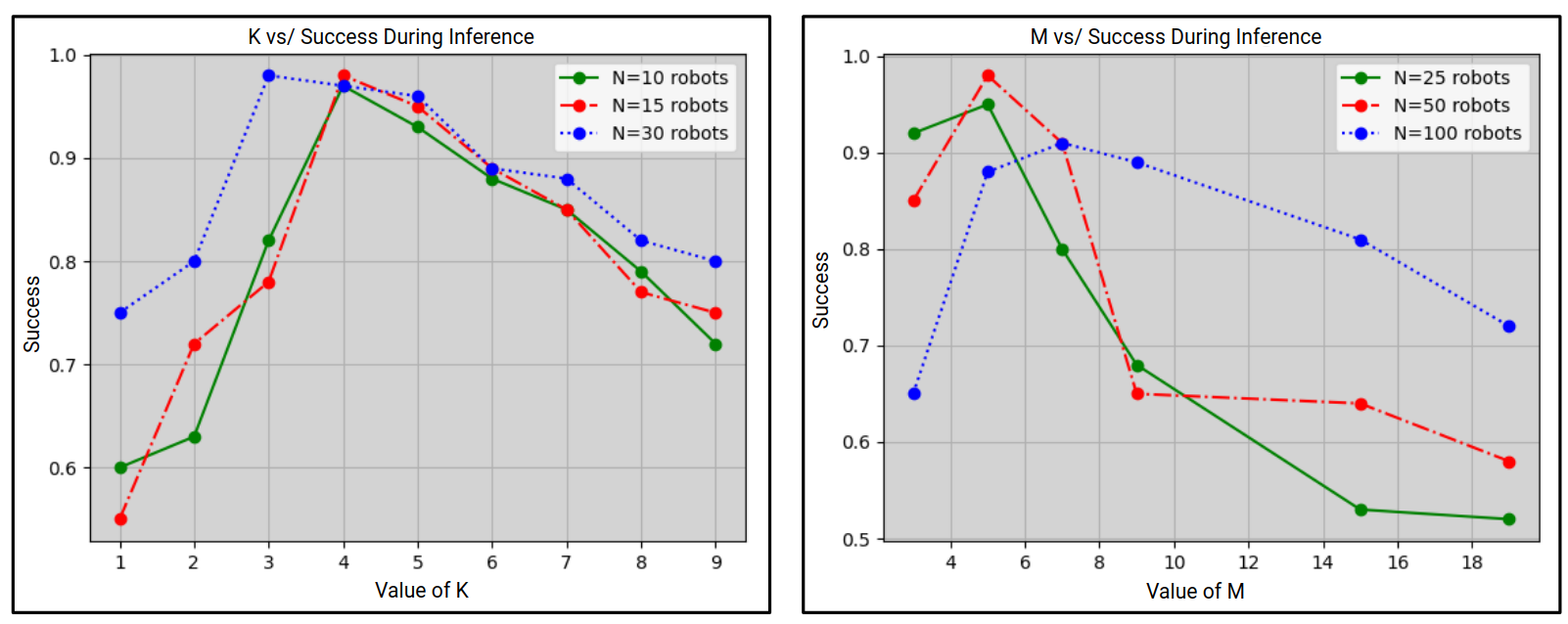}
  \caption{\textbf{Success v/s choice of K and M.} (L) During inference, we analyze effects of choices for K and M. 
  \label{fig:kandmplot}}
\end{figure}

We also study the effects of the choice of K and M on the performance of the learned models. We observe that as the number of hops, i.e $K$ each robot communicates with, the performance measured by looking at the success percentage (number of times all goals are covered without collisions during 100 runs) during inference increases. However, this increases only to a certain point after which there is a steep drop off in performance. This can be explained to the increase in input dimensionality at each node, thus making the problem more challenging. This result can be seen in Fig. \ref{fig:kandmplot} (Left). A similar effect can be seen when the value of $M$, i.e direct neighbors and goals that each robot senses. The increase in information only improves the learning performance to a certain point after which there is a drop off in performance as seen in Fig. \ref{fig:kandmplot}(Right).
\subsection{Comparison to Centralized Model Based Methods}
\begin{figure}[t!]
\begin{minipage}{.49\textwidth}
  \centering
  \includegraphics[width=\linewidth]{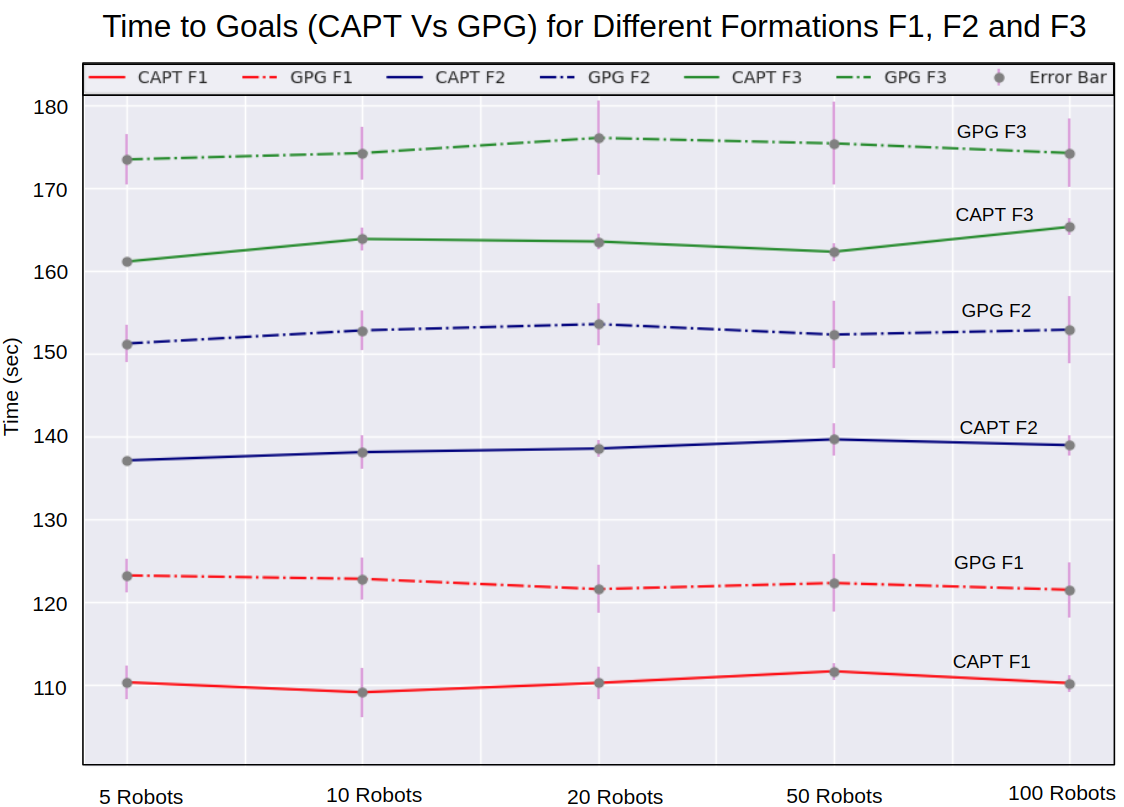}
  \caption{\textbf{Time to Goals and Planning Time} (Left) Time taken by CAPT to cover all goals v/s time taken by GPG to cover all goals when robots follow velocity controls. (Right) CAPT v/s GPG Planning Times.
  \label{fig:timefig}}
\end{minipage}
\begin{minipage}{.49\textwidth}
  \centering
  \includegraphics[width=\linewidth]{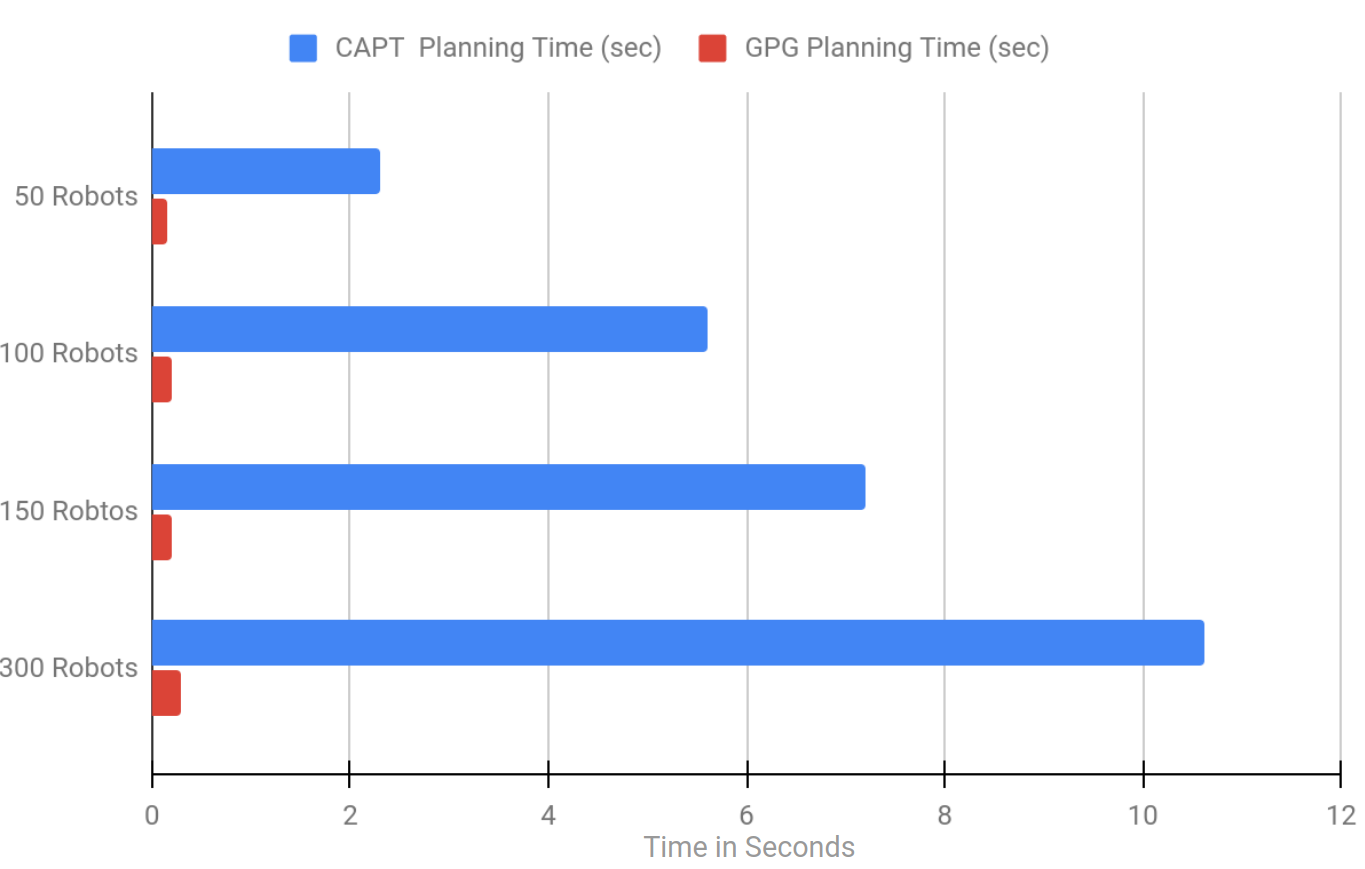}. 
\end{minipage}
\end{figure}

To further quantify the performance of GPG, we compare with a model based approach that uses centralized information, called concurrent assignment and planning (CAPT), proposed in \cite{turpin2014capt}. When used in an obstacle free environment, it guarantees collision free trajectories. We direct the reader to \cite{turpin2014capt} for more details about the CAPT algorithm. We set up three different formations F1, F2 and F3 similar to that in Fig \ref{fig:formationfig} (Center). On average the goals in F3 are further away from the starting positions of the robots than those in F2 and those in F2 are further away from goals in F1. In this work, we treat CAPT as the oracle and look to compare how well GPG performs when compared to this oracle. We use time to goals as a metric to evaluate GPG against this centralized oracle. Our results can be seen in Fig. \ref{fig:timefig} (Left). The key takeaway from this experiment is that decentralized inference using GPG, performs always within an $\epsilon$ margin (approximately 12-15 seconds) of the optimal solution and this margin remains more or less constant even if the goals are further away and if the number of robots are increased. However, GPG outshines CAPT when the planning times for both are compared. CAPT employs the hungarian algorithm to solve the task assignment and as a result has a time complexity of $\mathcal{O}(n^3)$ whereas GPG only requires a feedforward pass during inference. The planning times for both methods can be seen Fig.  \ref{fig:timefig} (Right).  It is important to note that in the original CAPT method, the optimal performance guarantees zero collisions. While the solution proposed in this paper is a learned solution with the advantage of being decentralized, we cannot guarantee collision free trajectories. Nevertheless, we analyze the number of collisions between robots. This performance can be seen in Table . We argue that while any learned method cannot guarantee collision free trajectories, it has been shown in the past that simple model based \textit{backup} policies can be used in conjunction with learned policies to empirically guarantee collision free trajectories\cite{khan2019learning}. To analyze this, we use velocity obstacles \cite{fiorini1998motion} as a backup policy that intervenes every time a collision between two robots executing GPG policies is imminent. We call this method \textbf{GPG+VO}. It is important to note that in order to prevent GPG+VO from producing degenerate solutions at the cost of collision free trajectories, we add a penalty term to the cost function every time VO is called during training. 
We analyze these results in Table \ref{tab:table1} for 30 robots over 3 line to circle formations similar to the one seen in Fig \ref{fig:formationfig} (Right). The formations R1 < R2 < R3 differ from each other in the size of the radius of the circle on which the goals are distributed. As in Fig \ref{fig:timefig} (Left) we observe that the difference between CAPT and GPG is the same even with different number of formations. In terms of number of collisions when the robots have more open space to operate, the number of collisions are almost reduced to zero with just GPG. When the room to operate is small as in R1 the GPG based policy tends to produce a higher number of collisions. The backup model based policy in GPG+VO is able to remove the number of collisions but results in an increase in time to cover goals. Using the backup model based velocity obstacles, increases the time required by an almost negligible amount when the robots have significant room to operate such as in R3 and at the same time offers a decentralized solution. Hence, we conclude that decentralized GPG is close in performance to the provably optimal centralized solution.

\begin{table}[]
\begin{tabular}{|l|l|l|l|l|l|l|}
\hline
\multirow{2}{*}{} & \multicolumn{2}{l|}{\textbf{CAPT}} & \multicolumn{2}{l|}{\textbf{GPG}} & \multicolumn{2}{l|}{\textbf{GPG+VO}} \\ \cline{2-7} 
 & T(sec) & \ C & T(sec) & \ C & T(sec) & \ C \\ \hline
\textbf{R1} & 68.27 & 0 & 85.31 & 11.4 & 94.9 & 0 \\ \hline
\textbf{R2} & 83.4 & 0 & 103.56 & 8.1 & 112.62 & 0 \\ \hline
\textbf{R3} & 98.16 & 0 & 122.15 & 2.3 & 124.15 & 0 \\ \hline
\end{tabular}
\caption{\textbf{CAPT vs GPG vs GPG+VO.} Total time (T(sec)) and  total number of collisions (C) during inference for 30 robots over 3 line to circle formations R1, R2 and R3 similar to the one seen in Fig \ref{fig:formationfig} (Right). R1 < R2 < R3 differ from each other in the size of the radius of the circle on which the goals are distributed. Averaged over 20 runs.}
\label{tab:table1}
\end{table}

\subsection{High Order Dynamics}
\label{subsec:expsairsim}
\begin{figure}[h!]
  \centering
  \includegraphics[width=\linewidth]{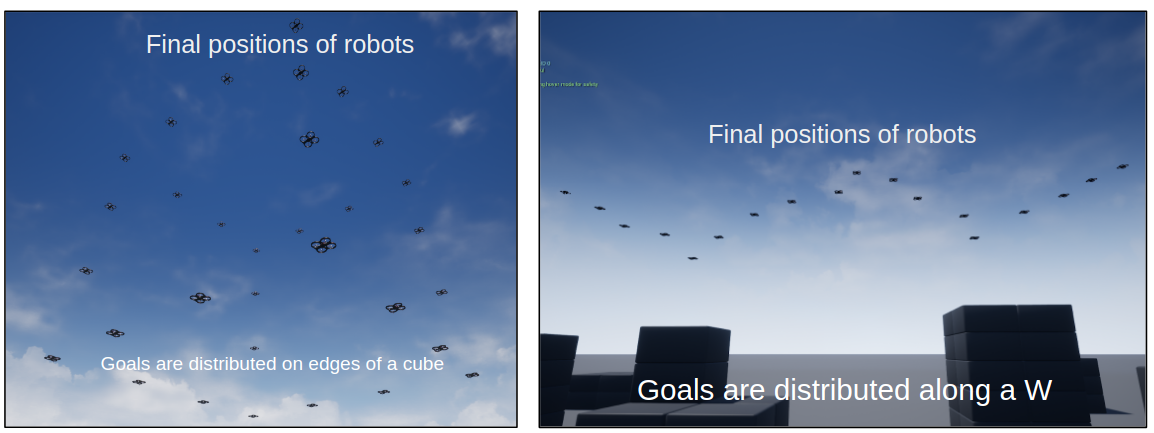}
  \caption{\textbf{Large Scale Unlabelled Motion Planning in AirSim using GPG.} Control to the training algorithm is handed after robots are at a certain altitude. (L) During inference, we use the trained filters to cover goals spread on the edges of a cube. (R) Goals to be covered are spread along a W. 
  \label{fig:airsimfig}}
\end{figure}

In the previous experiments, we modeled the robots as ideal point mass systems to study the feasibility of GNNs for the unlabeled motion planning problem. In this experiment we test our learned policies in the AirSim simulator which allows us to test our policies in the presence of higher order dynamics, slower control rates and latency in observations thus mimicking a real world robot swarm more realistically. In such a setting it is even more imperative to be able to train with a smaller number of robots and be able to transfer to a larger number without any additional training samples due to the increased computational complexity associated with training a large number of robots directly in the simulator. In this setting, the outputs from the policies are interpreted as accelerations and are converted to desired roll and pitch commands which are fed into the simulator. 
Further, since our proposed framework is model free, it can still be trained directly as before with only a sparse reward. Empirically, in Fig \ref{fig:airsimfig} and the attached video we observe our 
The action $\bbS_{nt}$ chosen by the robot gives the change in velocities. We execute the same paradigm as before. The unlabeled motion planning is trained for a small number of robots. Then, during inference time the filters learned for the small swarm are used across all the robots. A snapshot of the performance of our algorithm can be seen in Fig \ref{fig:airsimfig}.

\section{Conclusion}
In this paper, we look to achieve scalable decentralized solutions for the full unlabelled motion planning problem. In this work, we propose connecting the robots with a naive graph and utilize this graph structure to generate policies from local information by employing GNNs. We show that these policies can be transferred over to a larger number of robots and the solution computed is close to the solution computed by a centralized $\textit{oracle}$. One of the caveats of this paper is that it is assumed nearest robots can always be sensed which can potentially break down in a real world setting. Incorporating this is a potential avenue we leave for future work.





\bibliographystyle{IEEEtran}
\bibliography{root}

\end{document}